\pdfoutput=1
\documentclass[11pt]{article}
\usepackage[a4paper,margin=2.3cm]{geometry}
\usepackage[T1]{fontenc}
\usepackage{lmodern}
\usepackage{amsmath,amssymb}
\usepackage{graphicx}
\usepackage{booktabs}
\usepackage{xcolor}
\usepackage{caption}
\usepackage{setspace}
\usepackage[super,sort&compress,comma]{natbib}
\usepackage[hidelinks]{hyperref}
\newcommand{\accEnsAbs}{0.022}

\newcommand{\accEnsTwentyTwoAbs}{0.019}

\newcommand{\accGcAbs}{0.005}

\newcommand{\accHresAbs}{0.080}

\newcommand{\accHresTwentyTwoAbs}{0.078}
\newcommand{\accWinEns}{100}
\newcommand{\accWinEnsTwentyTwo}{100}

\newcommand{\accZfiveFive}{0.945}
\newcommand{\accZfiveTen}{0.671}
\newcommand{\bssPositiveDaysTwenty}{9.0}
\newcommand{\bssPositiveDaysTwentyTwo}{9.5}

\newcommand{\bssTFiveTwentyTwo}{0.48}

\newcommand{\bssTOneTwentyTwo}{0.70}

\newcommand{\bssTTenTwentyTwo}{0.24}

\newcommand{\bssZFiveTwentyTwo}{0.65}

\newcommand{\bssZOneTwentyTwo}{0.87}

\newcommand{\bssZTenTwentyTwo}{0.26}
\newcommand{\cardArchesTwenty}{72}
\newcommand{\cardEnsTwenty}{97}
\newcommand{\cardEnsTwentyTwo}{96}

\newcommand{\cardGcOpTwentyTwo}{69}

\newcommand{\cardNArchesTwenty}{190}

\newcommand{\cardNTwentyTwo}{570}
\newcommand{\growthEadyT}{0.79}
\newcommand{\growthEadyZ}{0.91}
\newcommand{\growthEq}{1.5}
\newcommand{\growthFiftyN}{4.2}
\newcommand{\growthInjZ}{-0.81}

\newcommand{\growthThirtyN}{2.6}

\newcommand{\inferEnsemble}{69}
\newcommand{\inferEnsembleCkpt}{23}
\newcommand{\inferSixty}{3.2}
\newcommand{\initsTwenty}{732}
\newcommand{\initsTwentyTwo}{700}
\newcommand{\injTropicsRatio}{3.8}

\newcommand{\membersTwentyTwo}{51}
\newcommand{\nArchesTwenty}{70}

\newcommand{\nNgcmTwenty}{76}
\newcommand{\nSaeInits}{121}
\newcommand{\params}{68.9}
\newcommand{\paramsS}{20.8}
\newcommand{\rmseDayOneEnsTeight}{0.77}
\newcommand{\rmseDayOneEnsTtwom}{0.85}
\newcommand{\rmseDayOneEnsUten}{0.81}
\newcommand{\rmseDayOneEnsZfive}{45.18}
\newcommand{\rmseDayOneGainTeight}{23}
\newcommand{\rmseDayOneGainTtwom}{28}
\newcommand{\rmseDayOneGainUten}{8}
\newcommand{\rmseDayOneGainZfive}{7}
\newcommand{\rmseDayOneTeight}{0.59}
\newcommand{\rmseDayOneTtwom}{0.61}
\newcommand{\rmseDayOneUten}{0.75}
\newcommand{\rmseDayOneZfive}{41.91}
\newcommand{\saeDead}{0}
\newcommand{\saeEv}{0.64}
\newcommand{\saeRtwoTBase}{0.090}
\newcommand{\saeRtwoTSae}{0.154}
\newcommand{\saeRtwoZBase}{0.012}
\newcommand{\saeRtwoZEady}{0.016}
\newcommand{\saeRtwoZSae}{0.046}
\newcommand{\saeTopFifty}{14}

\newcommand{\skillArchesTwenty}{-0.5}

\newcommand{\skillEnsQsevenTwentyTwo}{+7.8}

\newcommand{\skillEnsTeightTwentyTwo}{+10.9}

\newcommand{\skillEnsTtwomTwentyTwo}{+16.8}

\newcommand{\skillEnsTwenty}{+8.8}

\newcommand{\skillEnsTwentyTwo}{+8.3}
\newcommand{\skillEnsTwentyTwoAbs}{8.3}

\newcommand{\skillEnsUtenTwentyTwo}{+4.2}

\newcommand{\skillEnsZfiveTwentyTwo}{+5.1}

\newcommand{\skillFgnTwentyTwoAbs}{6.3}
\newcommand{\skillGcOpTwenty}{+5.2}

\newcommand{\skillGcOpTwentyTwo}{+2.2}

\newcommand{\skillGcTwentyAbs}{2.8}
\newcommand{\skillNgcmTwenty}{+2.4}

\newcommand{\skillSEnsTwenty}{+3.3}

\newcommand{\skillSEnsTwentyTwo}{+2.8}

\newcommand{\ssrTwenty}{0.99}
\newcommand{\ssrTwentyTwo}{0.99}
\newcommand{\trainGpuDays}{2.7}
\newcommand{\trainGpuHours}{65}
\newcommand{\winArchesTwenty}{49}

\newcommand{\winGcOpTwenty}{92}
\newcommand{\winGcOpTwentyTwo}{65}

\newcommand{\winNgcmTwenty}{79}
\newcommand{\tradMaxShift}{0.56}
\newcommand{\kappaLo}{0.72}
\newcommand{\kappaHi}{0.94}
\newcommand{\model}{Sonny-ENS}

\title{\model: demystifying machine-learning ensemble weather forecasting}
\author{Minjong Cheon\\[2pt]
\small Department of Computer Science, Korea National Open University, Seoul, Republic of Korea\\
\small \texttt{jmj2316@mail.knou.ac.kr}}
\date{}

\begin{document}
\maketitle

\begin{abstract}
\noindent
Every weather forecast is uncertain, and decisions from flood warnings to energy trading depend on how well that
uncertainty is quantified. Machine-learning (ML) ensembles now quantify it better than the best physics-based
ensemble, yet two things about them remain opaque: whether their skill requires hundreds to thousands of
accelerator-days of training, and how a network with a simple source of randomness produces a spread that follows the
flow. Here we introduce \model, an ensemble of three \params-million-parameter networks with Monte Carlo dropout,
trained on the continuous ranked probability score (CRPS) in \trainGpuDays{} A40 GPU-days. On 2022, it has a lower
CRPS than the European Centre for Medium-Range Weather Forecasts ensemble (ENS) on \cardEnsTwentyTwo\% of
\cardNTwentyTwo{} targets and an \skillEnsTwentyTwoAbs\% lower mean CRPS at lead times of 1--10 days. It outperforms
the operational GenCast on \cardGcOpTwentyTwo\% of the targets, and a 51-member, 15-day forecast takes \inferEnsemble{}
s on one GPU. Ablations identify the ingredients of this skill, and a Kolmogorov--Arnold sparse autoencoder shows that
dropout injects uncertainty mainly in the tropics, whereas the spread grows where the atmosphere is baroclinically
unstable (spatial correlation \growthEadyZ{} with the Eady growth rate). Skillful probabilistic weather forecasts are
therefore within reach of a single GPU server.
\end{abstract}

% ---------------------------------------------------------------------------------------------------------------
Small errors in a weather forecast do not grow everywhere alike: they amplify fastest where the atmosphere is
baroclinically unstable, along the mid-latitude storm tracks\cite{eady1949long,lindzen1980simple,hoskins1990storm}.
Forecast uncertainty therefore depends on the flow and has to be predicted together with the weather
itself\cite{palmer2000predicting,leutbecher2008ensemble}. Operational centers do so with ensembles designed around this
physics. The ENS of the European Centre for Medium-Range Weather Forecasts (ECMWF) integrates 50 perturbed copies of
its physics-based model, with initial perturbations from singular vectors, which target the fastest-growing errors, and
from ensemble data assimilation, and with stochastic perturbations of the physical
tendencies\cite{molteni1996ecmwf,buizza1999stochastic,palmer2019stochastic}; other centers use related
methods\cite{toth1997ensemble}. These ensembles are among the largest computations in operational
meteorology\cite{bauer2015quiet}.

ML models trained on reanalysis now produce deterministic medium-range forecasts that are more accurate than the best
physics-based deterministic system\cite{bi2023pangu,lam2023graphcast,chen2023fuxi,nguyen2023stormer,cheon2026sonny}.
Trained to minimize squared error, these models blur their forecasts as lead time grows and do not describe the range
of possible outcomes\cite{price2025gencast}. A second generation of ML models produces ensembles. GenCast, a
conditional diffusion model, outperformed ENS on 97.2\% of 1,320 targets\cite{price2025gencast}. NeuralGCM couples a
differentiable dynamical core with learned physics\cite{kochkov2024neuralgcm}; FGN, AIFS-CRPS and FuXi-ENS train
directly on proper scoring rules or variational objectives\cite{alet2025fgn,lang2026aifscrps,zhong2025fuxiens};
ArchesWeatherGen corrects a deterministic forecast with a generative model\cite{couairon2024archesgen}; and huge
ensembles of spherical Fourier neural operators sample thousands of members\cite{mahesh2024huge}.

These gains have been obtained with large computational budgets. As reported, GenCast was trained for 160 TPUv5-days,
FGN for 490 TPU-days, AIFS-CRPS for 896 H100-days and NeuralGCM ENS for 1,280 TPUv5e-days\cite{ucast2026}
(Extended Data Table~\ref{edt:cost}). Recent models lowered the cost: ArchesWeatherGen trains in 45 V100-days and U-Cast, an
ensemble of four 895-million-parameter U-Nets at 1.5$^\circ$, in 11.8 H200-days\cite{couairon2024archesgen,ucast2026}.
Training cost decides who can develop, retrain and adapt an ensemble system, how quickly design choices can be
tested, and how many members can be afforded when the system runs.

Machine learning offers inexpensive ways to turn a single network into a sampler. Monte Carlo (MC) dropout keeps
dropout active at inference and draws a new network for every sample\cite{gal2016dropout}, and deep ensembles combine
independently trained networks\cite{lakshminarayanan2017simple}; early neural-network weather ensembles built this
way were under-dispersive\cite{scher2021ensemble,garg2022wbprob,ucast2026}. U-Cast combined both with training on a proper
scoring rule\cite{gneiting2007strictly}, the CRPS, and with four 895-million-parameter U-Nets matched the skill of
GenCast at 1.5$^\circ$\cite{ucast2026}. Whether these mechanisms still work when each network is more than ten times
smaller and trained on a few GPUs of the previous generation has not been established, and it is not known how a
network whose only source of randomness is dropout produces a spread that depends on the flow.

Here we present \model{} and use it to answer both questions. \model{} extends the deterministic model
Sonny\cite{cheon2026sonny} with MC dropout, CRPS training and a deep ensemble of three rollout-trained models; with
\params{} million parameters per network and \trainGpuDays{} A40-days of training, it outperforms ENS on
\cardEnsTwentyTwo\% of the 2022 WeatherBench~2 (WB2)\cite{rasp2024wb2} scorecard. We compare it with every ensemble in
WB2 on two test years and examine forecasts of a record heatwave and a typhoon. Ablations then show which ingredients
carry the skill, and a Kolmogorov--Arnold sparse autoencoder
(KAN-SAE)\cite{cheon2026kansae,liu2024kan,cunningham2023sparse} shows where the ensemble spread is created and where
it grows.

% ---------------------------------------------------------------------------------------------------------------
\section*{\model}

\begin{figure}[tbp]
\centering
\includegraphics[width=\textwidth]{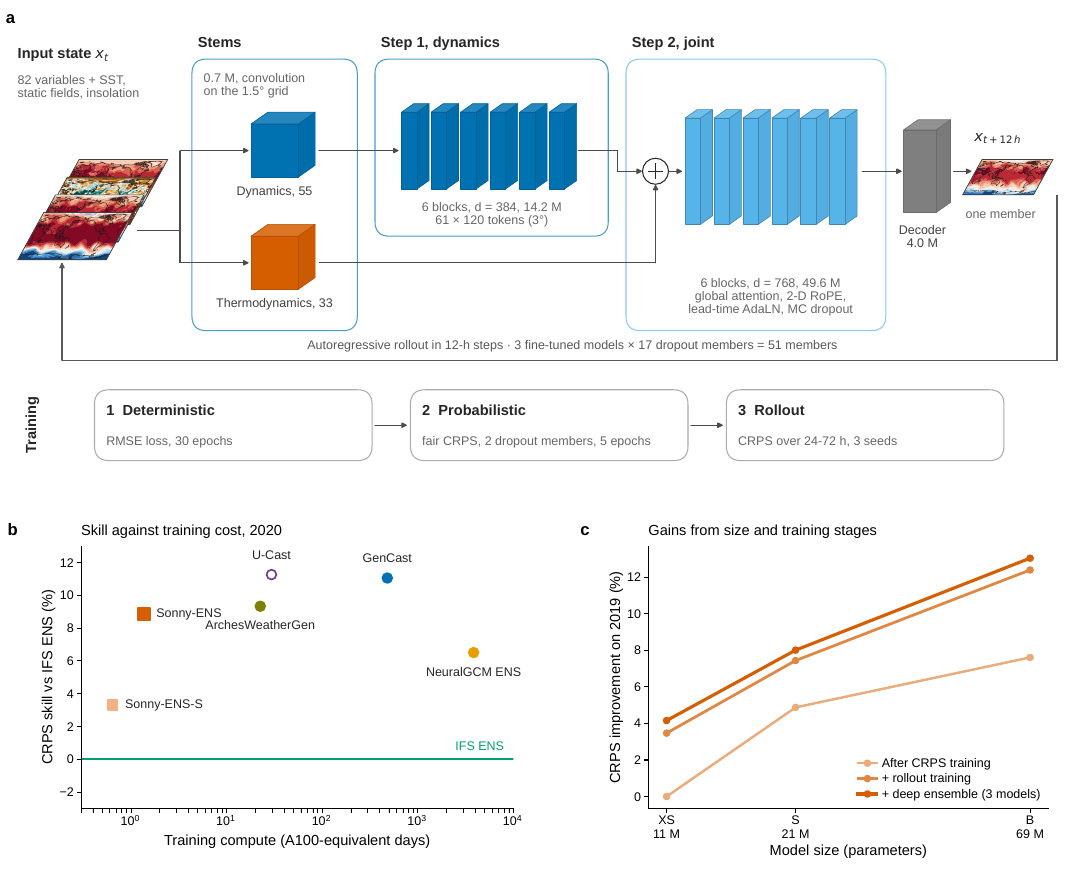}
\caption{\textbf{| A \params-million-parameter ensemble trained in under three A40-days.} \textbf{a}, The model. The
state at time $t$ enters two full-resolution convolutional stems with spherical padding (dynamics: geopotential,
winds, vertical velocity, MSLP and 10-m winds; thermodynamics: temperature, humidity and T2m, with SST, static
fields and insolation). Step~1 runs six transformer blocks on the dynamical tokens (61\,$\times$\,120, 3$^\circ$);
step~2 adds the thermodynamical tokens and runs six wider blocks with global attention, 2-D rotary positions,
lead-time AdaLN and MC dropout; a decoder returns the 12-h tendency. Maps: the analysis at 12\,UTC 14 July 2022 and
one member 12\,h later. Training has three stages (bottom). \textbf{b}, Mean CRPS skill against IFS ENS on 2020
(seven headline variables, days 1--10 every 12\,h) against training compute in A100-equivalent days (conversion
factors in Extended Data Table~\ref{edt:cost}). U-Cast is placed from its reported skill relative to GenCast.
\textbf{c}, CRPS improvement on 2019 (held out) relative to the smallest model after CRPS training, for three model
sizes and each training stage.}
\label{fig:overview}
\end{figure}

\model{} maps the global atmospheric state at time $t$ on a 1.5$^\circ$ latitude--longitude grid
(121\,$\times$\,240 points) to the state at $t+12$\,h and produces 15-day forecasts by feeding its output back as input
(Fig.~\ref{fig:overview}a). The state has 82 variables: six upper-air variables on 13 pressure levels and four surface
variables. Sea-surface temperature (SST), two static fields and top-of-atmosphere insolation enter as forcings. The
network has three parts. Two convolutional stems on the full-resolution grid encode the dynamical variables
(geopotential, winds, vertical velocity, mean sea-level pressure and 10-m winds) and the thermodynamical variables
(temperature, humidity and 2-m temperature, together with the forcings), with padding that follows the sphere across
the dateline and the poles, into tokens on a 3$^\circ$ grid (61\,$\times$\,120). A first stack of six transformer
blocks processes the dynamical tokens alone. A second stack of six wider blocks processes them jointly with the
thermodynamical tokens, using global attention with two-dimensional rotary position
embedding\cite{su2024roformer} and lead-time conditioning\cite{peebles2023dit}. A convolutional decoder returns the
12-h tendency on the 1.5$^\circ$ grid. This two-step design, in which the dynamics is resolved before thermodynamics
is added, follows Sonny\cite{cheon2026sonny}.

\model{} is stochastic because dropout stays active during forecasts\cite{gal2016dropout}, as in
U-Cast\cite{ucast2026}: each member draws new
dropout masks inside every transformer block at every 12-h step, so one trained network samples a distribution of
trajectories. Training has three stages (Fig.~\ref{fig:overview}a, bottom; Methods). The network is first trained
deterministically on 1979--2018 with a per-variable root-mean-square error (RMSE) loss\cite{otter2026}. It is then
trained for five epochs with the fair CRPS\cite{gneiting2007strictly,zamo2018estimation} computed over two dropout
members, which turns the dropout into a model of forecast uncertainty. Finally, three copies are fine-tuned with the
CRPS on autoregressive rollouts of 24--72\,h, with different random seeds and learning rates. The ensemble pools the
three models\cite{lakshminarayanan2017simple} with 17 dropout members each, giving 51 members, about the size of
ENS. Training takes \trainGpuHours{} A40 GPU-hours (\trainGpuDays{} A40-days) in total, and a 51-member, 15-day
forecast takes about \inferEnsemble{} s on one A40, or \inferEnsembleCkpt{} s with the three models on three GPUs
(Extended Data Table~\ref{edt:cost}). Fig.~\ref{fig:overview}b places \model{} among published ML
ensembles by training cost and skill, and Fig.~\ref{fig:overview}c shows that skill increases with model size and with
each training stage.

We evaluate \model{} with the WB2 protocol\cite{rasp2024wb2}: forecasts and the ERA5 reanalysis\cite{hersbach2020era5}
as truth on the 1.5$^\circ$ grid, initial times at 00 and 12\,UTC and lead times from 12\,h to 15 days. \model{} starts
from ERA5. All design decisions were made on 2019, which was withheld from training. In 2022 (\initsTwentyTwo{}
initial times) we compare with the operational systems in WB2: ENS, the operational GenCast and FGN, all initialized
from operational analyses. In 2020 (\initsTwenty{} initial times) we compare with GenCast initialized from ERA5,
NeuralGCM ENS and ArchesWeatherGen. Unless stated otherwise, the headline score is the CRPS skill,
$100\,(1-\mathrm{CRPS}/\mathrm{CRPS}_\mathrm{ref})$, averaged over seven variables (Z500, T850, Q700, U850, T2m, U10
and MSLP) and over lead times of 1--10 days every 12\,h (133 targets)\cite{otter2026}. Our metric code agrees with the
official WB2 code to a relative difference below $10^{-8}$ (Supplementary Table~\ref{st:wb2}).

% ---------------------------------------------------------------------------------------------------------------
\section*{Ensemble skill}

\begin{figure}[tbp]
\centering
\includegraphics[width=\textwidth]{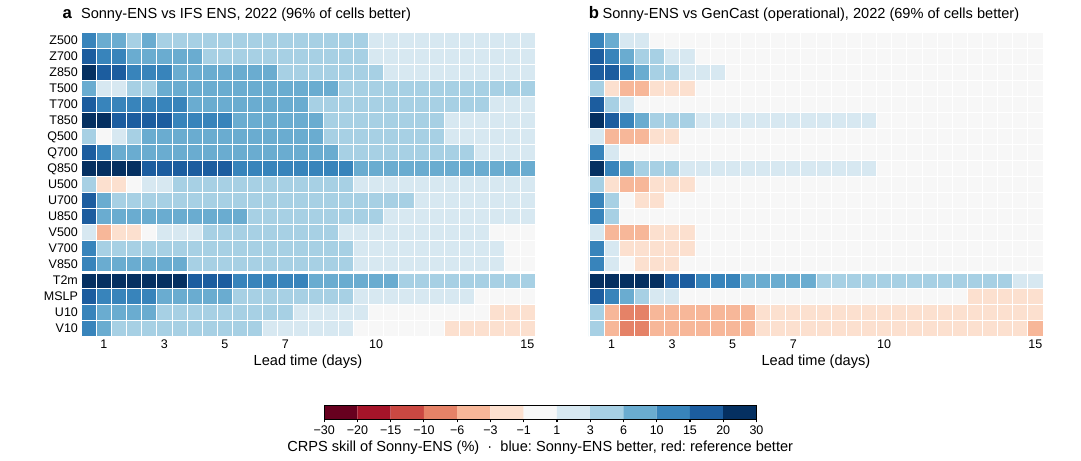}
\caption{\textbf{| \model{} outperforms ENS almost everywhere in 2022.} CRPS skill,
$100\,(1-\mathrm{CRPS}_\text{\model}/\mathrm{CRPS}_\mathrm{ref})$, by variable and lead time (0.5--15 days) on the
2022 test year (\initsTwentyTwo{} initial times; \membersTwentyTwo{} members) against \textbf{a}, IFS ENS and
\textbf{b}, GenCast (operational).
Blue: \model{} better; red: reference better.}
\label{fig:scorecards}
\end{figure}

On 2022, \model{} has a lower CRPS than ENS for \cardEnsTwentyTwo\% of the \cardNTwentyTwo{} combinations of 19
variables and 30 lead times (Fig.~\ref{fig:scorecards}a) and for all 133 headline targets, with a mean CRPS skill of
$\skillEnsTwentyTwo$\%. The improvement is largest for temperature and humidity near the surface and in the lower
troposphere (T2m $\skillEnsTtwomTwentyTwo$\%, T850 $\skillEnsTeightTwentyTwo$\%, Q700 $\skillEnsQsevenTwentyTwo$\%)
and smallest for geopotential and near-surface wind (Z500 $\skillEnsZfiveTwentyTwo$\%, U10
$\skillEnsUtenTwentyTwo$\%). Against the operational GenCast, which runs at 0.25$^\circ$, \model{} is better on
\cardGcOpTwentyTwo\% of the scorecard (Fig.~\ref{fig:scorecards}b) and by $\skillGcOpTwentyTwo$\% on the headline
targets, \winGcOpTwentyTwo\% of which it wins. FGN, trained with 490 TPU-days, has a \skillFgnTwentyTwoAbs\% lower
headline CRPS than \model{} (Extended Data Fig.~\ref{edf:skill22}). Starting from the operational HRES analysis instead of ERA5 raises
the day-1 CRPS of \model{} by 8.8--13.9\% and by less than 1.5\% after day 10 (Extended Data Fig.~\ref{edf:init}).

\begin{figure}[tbp]
\centering
\includegraphics[width=\textwidth]{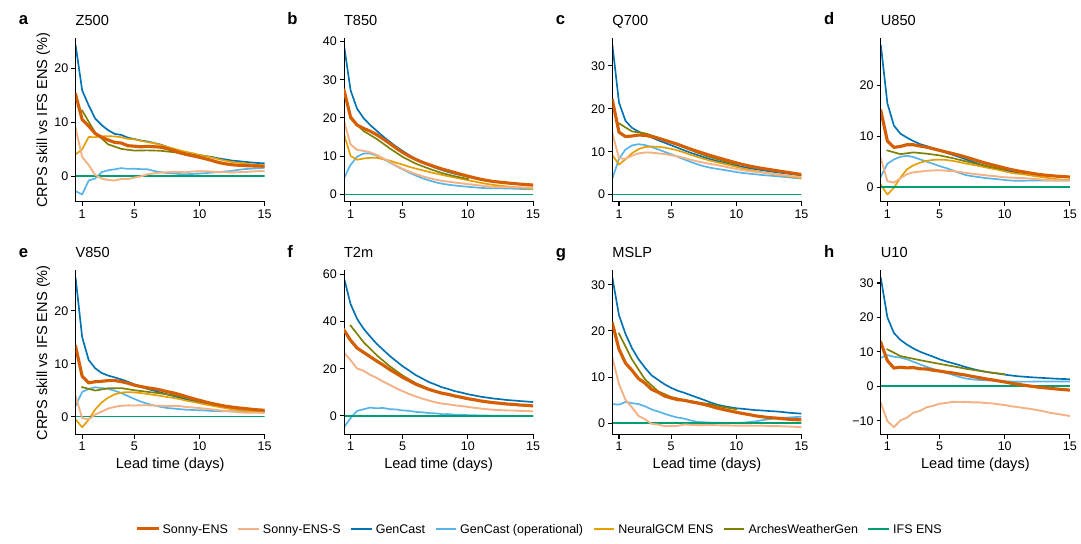}
\caption{\textbf{| Skill against ENS across lead times.} CRPS skill relative to IFS ENS (zero line) on 2020 for
\textbf{a--h}, Z500, T850, Q700, U850, V850, T2m, MSLP and U10, with every WB2 ensemble scored that year.}
\label{fig:skill}
\end{figure}

The 2020 test year reproduces the result against ENS ($\skillEnsTwenty$\%, all 133 targets, \cardEnsTwenty\% of the
scorecard) and against the operational GenCast ($\skillGcOpTwenty$\%, \winGcOpTwenty\% of targets)
(Fig.~\ref{fig:skill}, Extended Data Table~\ref{edt:scores} and Supplementary Figs.~\ref{sf:card20} and
\ref{sf:abs}). \model{} also outperforms NeuralGCM ENS
($\skillNgcmTwenty$\% over the \nNgcmTwenty{} targets it provides, \winNgcmTwenty\% of them), is level with
ArchesWeatherGen on the headline targets ($\skillArchesTwenty$\%, \winArchesTwenty\% of \nArchesTwenty{}) and better
on \cardArchesTwenty\% of its \cardNArchesTwenty-cell scorecard, including every upper-air temperature and wind cell
(Supplementary Fig.~\ref{sf:card20}b), and is within
\skillGcTwentyAbs\% of GenCast initialized from ERA5, which is trained at 0.25$^\circ$ with more than 100 times the
training compute (Fig.~\ref{fig:overview}b). Averaged over the five upper-air variables used to compare ArchesWeatherGen
and U-Cast\cite{couairon2024archesgen,ucast2026} (Z500, T850, Q700, U850 and V850), the CRPS skill of \model{} exceeds
those of ArchesWeatherGen, NeuralGCM ENS and the operational GenCast at every lead time, and that of GenCast from day 6
to day 11.5 (Extended Data Fig.~\ref{edf:summary}a). The smaller variant \model-S (\paramsS{} million parameters, 1.3
A40-days) also outperforms ENS in both years ($\skillSEnsTwentyTwo$\% in 2022 and $\skillSEnsTwenty$\% in 2020).

% ---------------------------------------------------------------------------------------------------------------
\section*{Ensemble-mean accuracy and calibration}

\begin{figure}[tbp]
\centering
\includegraphics[width=\textwidth]{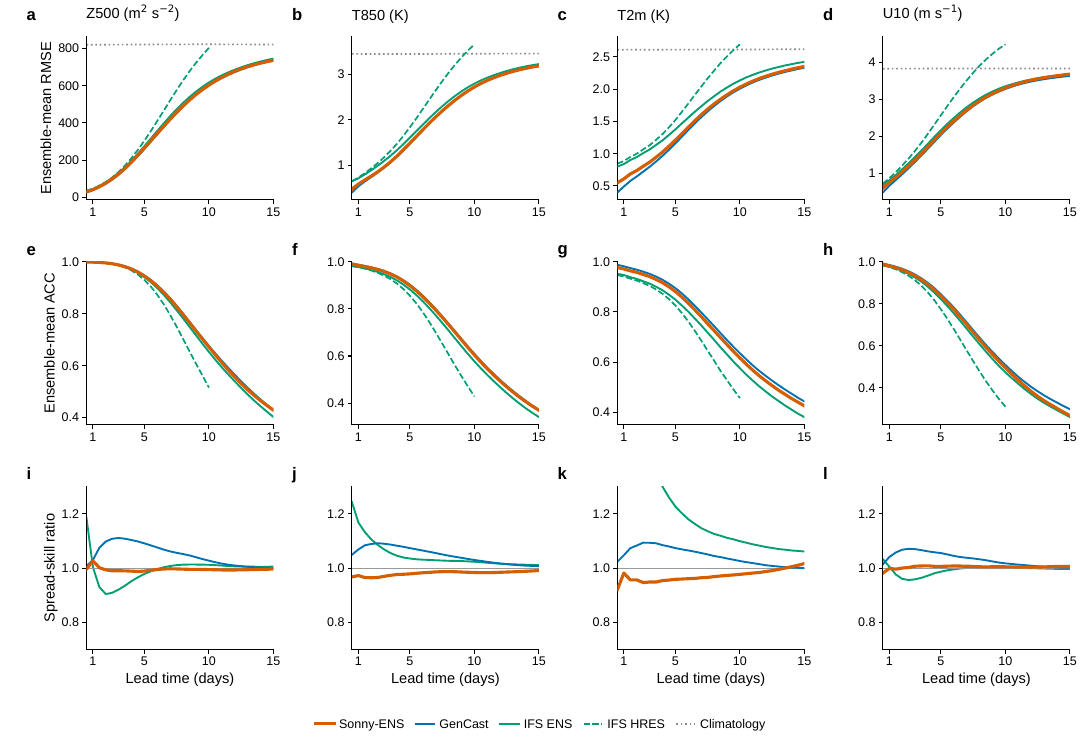}
\caption{\textbf{| Ensemble-mean accuracy and calibration.} 2020; Z500, T850, T2m and U10. \textbf{a--d},
Ensemble-mean RMSE (corrected for finite ensemble size for every ensemble) with IFS HRES and climatology.
\textbf{e--h}, Ensemble-mean anomaly correlation (WB2 ERA5 climatology). \textbf{i--l}, Spread--skill ratio
(corrected); 1 is perfect calibration.}
\label{fig:mean}
\end{figure}

A useful ensemble needs an accurate mean and a spread that matches the error of that mean\cite{fortin2014spread}. In
2020 the ensemble-mean RMSE of \model{} is below that of ENS for Z500, T850, T2m, U10 and MSLP at days 1, 3, 5 and 10
(Fig.~\ref{fig:mean}a--d and Extended Data Table~\ref{edt:scores}b). At day 1 in 2022 it is \rmseDayOneGainZfive\% lower than that of
ENS for Z500 (\rmseDayOneZfive{} against \rmseDayOneEnsZfive{} m$^2$\,s$^{-2}$), \rmseDayOneGainTeight\% lower for
T850 (\rmseDayOneTeight{} against \rmseDayOneEnsTeight{} K), \rmseDayOneGainTtwom\% lower for T2m
(\rmseDayOneTtwom{} against \rmseDayOneEnsTtwom{} K) and \rmseDayOneGainUten\% lower for U10 (\rmseDayOneUten{}
against \rmseDayOneEnsUten{} m\,s$^{-1}$). Averaged over the headline targets in 2020, the anomaly correlation
coefficient (ACC) of the ensemble mean exceeds that of ENS by \accEnsAbs{} (higher on \accWinEns\% of targets) and
that of the deterministic IFS HRES by \accHresAbs{} (all targets), and is \accGcAbs{} below that of GenCast; the Z500
ACC is \accZfiveFive{} at day 5 and \accZfiveTen{} at day 10 (Fig.~\ref{fig:mean}e--h). In 2022 the ACC exceeds that
of ENS by \accEnsTwentyTwoAbs{} (\accWinEnsTwentyTwo\% of targets) and that of HRES by \accHresTwentyTwoAbs.

On the same five upper-air variables, the ensemble mean of \model{} is more accurate than those of ArchesWeatherGen,
NeuralGCM ENS and the operational GenCast, the Stormer ensemble mean and the deterministic IFS HRES and Pangu-Weather at
every lead time, and than GraphCast from day 4 (Extended Data Fig.~\ref{edf:summary}b).

The ensemble is calibrated. The spread--skill ratio, corrected for ensemble size, averages \ssrTwenty{} over the
headline variables and days 1--10 in 2020 and \ssrTwentyTwo{} in 2022, against an ideal value of 1
(Fig.~\ref{fig:mean}i--l). Over the five upper-air variables it stays between 0.987 and 0.996 at every lead time, the
closest to 1 of all ensembles in WB2: its mean departure from 1 is 0.007, against 0.015 to 0.106 for the others
(Extended Data Fig.~\ref{edf:summary}c). Individual members keep realistic small-scale variability: their zonal power spectra for
T850, Q700 and T2m stay within 2.5\% of ERA5 up to day 10 (Extended Data Fig.~\ref{edf:spec}).

% ---------------------------------------------------------------------------------------------------------------
\section*{High-impact weather}

\begin{figure}[tbp]
\centering
\includegraphics[width=\textwidth]{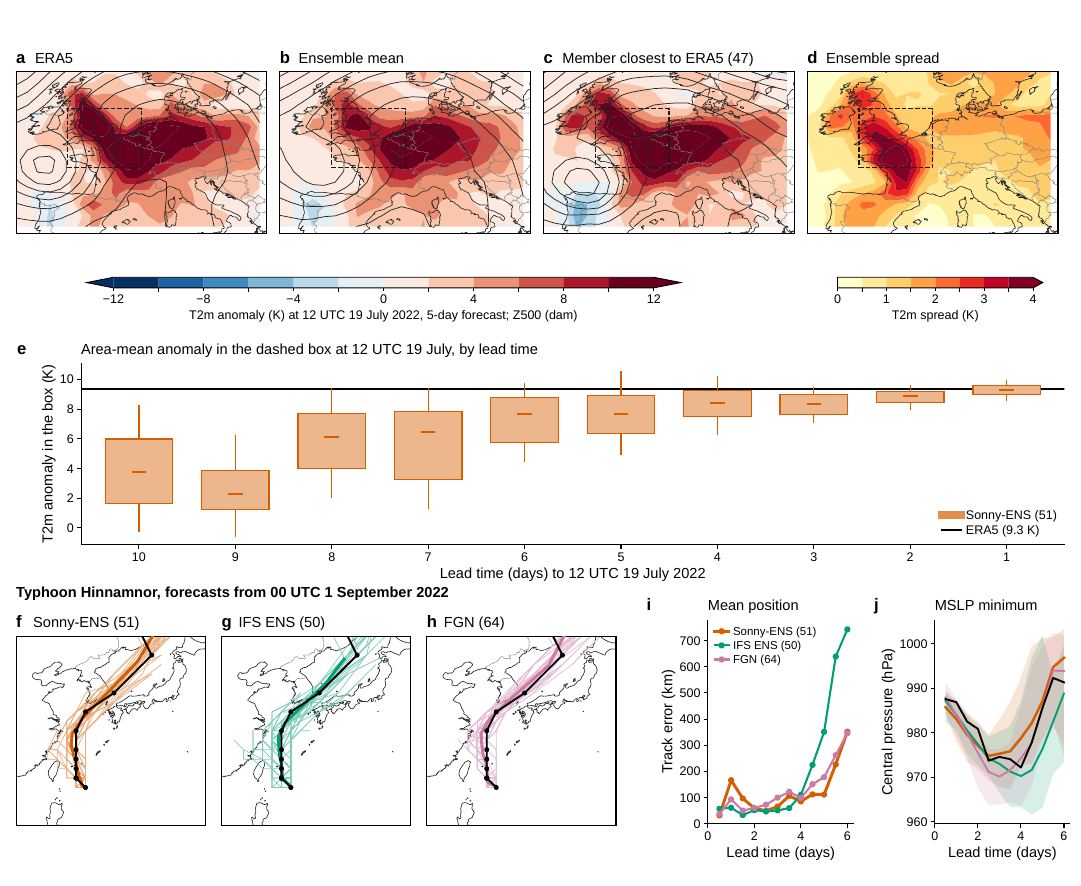}
\caption{\textbf{| The 2022 European heatwave and Typhoon Hinnamnor.} \textbf{a--d}, \model{} (51 members): T2m
anomaly at 12\,UTC 19 July 2022 from the forecast started 5 days earlier: ERA5, ensemble mean, the member closest to
ERA5 in the dashed box (England and northern France) and ensemble spread; contours, Z500 (dam). \textbf{e}, Area-mean
T2m anomaly in the box at that time from forecasts started 1--10 days earlier (boxes, quartiles of the members;
whiskers, 5--95\%; bars, median) and ERA5 (line). \textbf{f--h}, Tracks of the MSLP minimum from 00\,UTC 1 September
2022 for \model{} (51 members), IFS ENS (50) and FGN (64): members (thin), ensemble-mean position (thick) and ERA5
(black, 12-hourly dots). \textbf{i}, Distance between the ensemble-mean position and ERA5. \textbf{j}, Central
pressure: median (lines) and 5--95\% (shading) of the members, and ERA5 (black). All forecasts on the 1.5$^\circ$
grid; \model{} starts from ERA5, IFS ENS and FGN from the operational analysis.}
\label{fig:cases}
\end{figure}

On 19 July 2022 temperatures in England exceeded 40\,$^\circ$C for the first time on record. Started five days earlier,
the 51-member \model{} forecasts the heat plume over England, France and the Low Countries and the cut-off low to its
west (Fig.~\ref{fig:cases}a--c). Over a box covering England and northern France, the ensemble-mean T2m anomaly at
that time is 7.7\,K against 9.35\,K in ERA5, and the members span 4.1--12.1\,K. The spread is largest over England and
western France, at the western edge of the plume (Fig.~\ref{fig:cases}d). From eight days ahead the ensemble median of
the box anomaly exceeds 6\,K, from four days ahead it is within 1\,K of ERA5, and ERA5 lies within the 5--95\% range of
the members at every lead time up to eight days (Fig.~\ref{fig:cases}e). IFS ENS and FGN forecasts of this event are
compared in Supplementary Fig.~\ref{sf:hw}.

For Typhoon Hinnamnor, started at 00\,UTC 1 September 2022, the members of \model, IFS ENS and FGN all follow the
northward turn of the storm toward the Korean Peninsula and Japan (Fig.~\ref{fig:cases}f--h). On days 4, 5 and 6 the
ensemble-mean position of \model{} lies 86, 112 and 352\,km from ERA5, against 110, 351 and 743\,km for IFS ENS and
99, 178 and 346\,km for FGN; on days 1--3 all three stay within 170\,km, about one 1.5$^\circ$ grid step
(Fig.~\ref{fig:cases}i). The median central pressure of \model{} differs from ERA5 by at most 4\,hPa except on day 4,
the deepest stage, and on day 6, when it is 7 and 6\,hPa higher (Fig.~\ref{fig:cases}j).

Over the whole 2022 test year, \model{} also predicts the probability of extremes, defined per grid point and calendar
month by the 5th and 95th percentiles of ERA5 in 1990--2019 (Extended Data Fig.~\ref{edf:ext}). The Brier skill
score against the climatological 5\% forecast for Z500 above its 95th percentile is \bssZOneTwentyTwo{} at day 1,
\bssZFiveTwentyTwo{} at day 5 and \bssZTenTwentyTwo{} at day 10, and for T2m \bssTOneTwentyTwo, \bssTFiveTwentyTwo{}
and \bssTTenTwentyTwo. For Z500, T2m and wind speed at 850\,hPa and 10\,m, warm and cold or strong and weak extremes
alike, the score stays positive up to day \bssPositiveDaysTwentyTwo{} (2020: day \bssPositiveDaysTwenty).

% ---------------------------------------------------------------------------------------------------------------
\section*{Spread grows where the atmosphere is baroclinically unstable}

How does a network whose only source of randomness is dropout produce a spread that is calibrated and depends on the
flow? We trained a KAN-SAE\cite{cheon2026kansae} with 2,048 features on 1.2 million token activations of the last transformer block of one
\model{} model (768 dimensions; Methods). The activations of this layer are dominated by a few massive
dimensions\cite{sun2024massive}: one dimension carries 59\% of the variance and five carry 95\%. After each dimension
is standardized, the KAN-SAE explains \saeEv{} of the variance on held-out tokens, and all its features are active.
We then ran one model with eight dropout members from \nSaeInits{} initial times in 2019 and measured, for each
3$^\circ$ token, the member-to-member variance of the KAN-SAE features at the first step (the uncertainty that
dropout injects) and the growth of the ensemble spread over the next three days.

\begin{figure}[tbp]
\centering
\includegraphics[width=\textwidth]{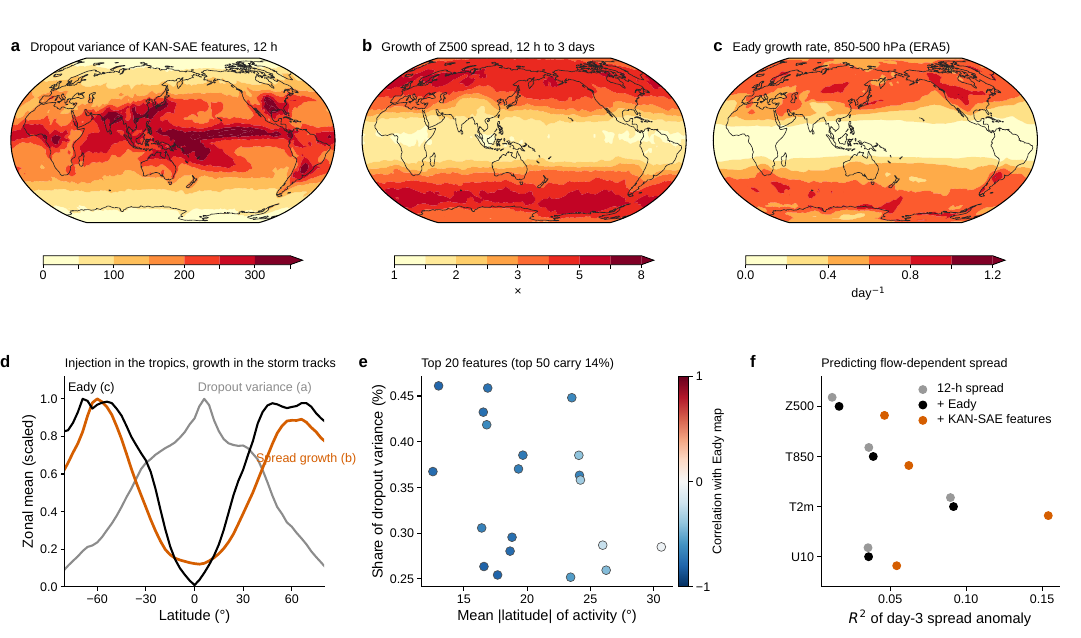}
\caption{\textbf{| \model{} injects uncertainty in the tropics and grows it in the storm tracks.} KAN-SAE analysis of
one \model{} model on 2019 (\nSaeInits{} initial times, 8 MC-dropout members, 3 days). \textbf{a}, Member-to-member
(dropout) variance of the KAN-SAE features of the last transformer block at the first step, summed over features
(standardized units). \textbf{b}, Growth of the Z500 ensemble spread from 12\,h to 3 days. \textbf{c}, Eady growth
rate of the 850--500-hPa layer from ERA5. The growth map follows the Eady map ($r=\growthEadyZ$) and not the
injection map ($r=\growthInjZ$). \textbf{d}, Zonal means of a--c, each scaled by its maximum. \textbf{e}, The 20
features carrying most dropout variance (the top 50 carry \saeTopFifty\%): share against mean latitude of activity,
colored by the spatial correlation of their dropout-variance map with the Eady map. \textbf{f}, Out-of-sample $R^2$
of the day-3 log-spread anomaly (odd initial times for fitting, even ones for testing) from the 12-h spread, adding
the Eady growth rate, and adding 100 KAN-SAE features.}
\label{fig:spread}
\end{figure}

Dropout injects uncertainty mainly in the tropics: the injected variance is \injTropicsRatio{} times larger within
20$^\circ$ of the equator than poleward of 45$^\circ$, with maxima over the Indo-Pacific warm pool, the intertropical
convergence zones and South and East Asia (Fig.~\ref{fig:spread}a). The spread does not grow where it is injected.
Over three days the Z500 spread grows by a factor of \growthFiftyN{} near 50$^\circ$N, \growthThirtyN{} near
30$^\circ$N and \growthEq{} at the equator (Fig.~\ref{fig:spread}b). The growth map is spatially correlated with the
Eady growth rate of the 850--500-hPa layer in ERA5\cite{eady1949long,lindzen1980simple} ($r=\growthEadyZ$; T850,
$r=\growthEadyT$) and anticorrelated with the injection map ($r=\growthInjZ$) (Fig.~\ref{fig:spread}b--d). \model{}
therefore amplifies its spread in the storm tracks, where baroclinic instability amplifies small
perturbations\cite{hoskins1990storm} and where the leading perturbations of physics-based ensembles
grow\cite{leutbecher2008ensemble}. No part of the training specifies this pattern.

The injection is spread over many features: the 50 features with the largest dropout variance carry \saeTopFifty\% of
it. The 20 largest are active at mean absolute latitudes of 13--31$^\circ$, and their dropout-variance maps are
anticorrelated with the Eady map (Fig.~\ref{fig:spread}e). The features also carry information on where the spread
departs from its usual pattern. In out-of-sample predictions of the day-3 log-spread anomaly, the 12-h spread explains
$R^2=\saeRtwoZBase$ for Z500; adding the Eady growth rate raises this to \saeRtwoZEady{} and adding 100 KAN-SAE
features to \saeRtwoZSae{} (T2m, \saeRtwoTBase{} to \saeRtwoTSae{}; Fig.~\ref{fig:spread}f).

% ---------------------------------------------------------------------------------------------------------------
\section*{Training cost and the ingredients of skill}

\model{} uses less training compute than any other ML ensemble we compare with (Extended Data Table~\ref{edt:cost}):
\trainGpuDays{} A40-days in total, against 11.8 H200-days for U-Cast, 45 V100-days for ArchesWeatherGen and 160 or more
accelerator-days for the 0.25$^\circ$ models. With the conversion factors of ArchesWeatherGen and
U-Cast\cite{couairon2024archesgen,ucast2026}, this is 1.4 A100-equivalent days, against 22.5 for ArchesWeatherGen and
29.5 for U-Cast. One member advances 60 steps (30 days) in \inferSixty{} s on an A40.
Ablations on 2019 show where the skill comes from (Extended Data Fig.~\ref{edf:abl}). In the deterministic design, measured by
the RMSE of a small model, replacing the Muon optimizer\cite{liu2025muon} with AdamW\cite{loshchilov2019adamw} costs
3.3\%, two input states instead of one 1.5\%, the mean squared error instead of the RMSE loss 1.2\%, removing
insolation 0.4\% and padding in longitude only 0.1\%. In the probabilistic pipeline, measured by the CRPS, the
\paramsS-million-parameter model is 5.2\% worse than the \params-million-parameter model trained with the same
recipe, omitting rollout fine-tuning costs 2.8\%, a single model instead of three 0.8\%, a weight of 0.1 instead of
1.0 on the 10-m winds during rollout training 0.6\% and holding the initial SST instead of its anomaly 0.2\%. Rollout fine-tuning also removes a drift of CRPS training in the tropics, where the Z500
ensemble-mean RMSE after CRPS training alone is 146\% higher at day 15 (Extended Data Fig.~\ref{edf:drift}).

% ---------------------------------------------------------------------------------------------------------------
\section*{Discussion}

\model{} shows that an ML ensemble trained in under three days on data-center GPUs of the previous generation can
outperform the leading operational ensemble on nearly every variable and lead time, and the operational GenCast on
most of them. Its components are individually simple and known: MC dropout and deep ensembles trained on the CRPS,
as in U-Cast\cite{ucast2026}, and the Muon optimizer. With 13 times fewer parameters per network than U-Cast, the
skill comes from combining them with a compact two-step transformer and rollout training on the CRPS, and each of
these contributes measurably (Extended Data Fig.~\ref{edf:abl}).

The KAN-SAE analysis gives a physical account of how this cheap stochasticity becomes a useful ensemble. Dropout
perturbs the model mostly in the tropics, as stochastic physics in operational ensembles perturbs mostly the
tendencies of convection-dominated regions\cite{buizza1999stochastic,palmer2019stochastic}, and the learned dynamics
amplifies those perturbations where the flow is baroclinically unstable. Sparse autoencoders, developed for language models\cite{cunningham2023sparse,gao2024scaling} and used to find
heatwave and typhoon features in the deterministic Sonny\cite{cheon2026kansae}, can thus test whether an ML weather model reproduces the error growth
of the atmosphere.

The measured limits of \model{} define its current position. GenCast initialized from ERA5 has a \skillGcTwentyAbs\%
lower headline CRPS in 2020 and FGN a \skillFgnTwentyTwoAbs\% lower one in 2022, both at 0.25$^\circ$ and with
more than 100 times the training compute. The members carry 31\% more power than ERA5 at the smallest resolved scales of Z500 and
U10 at day 10 (Extended Data Fig.~\ref{edf:spec}), and starting from an operational analysis rather than ERA5 raises the day-1 CRPS
by 8.8--13.9\% (Extended Data Fig.~\ref{edf:init}). Applying the same recipe at 0.25$^\circ$ and with operational initial states
is the next step.

% ---------------------------------------------------------------------------------------------------------------
\section*{Methods}

\subsection*{Data}
We use ERA5\cite{hersbach2020era5} on the 1.5$^\circ$ WB2 grid\cite{rasp2024wb2} (121 latitudes including both poles,
240 longitudes) every 12\,h, at 00 and 12\,UTC. ERA5 assimilates observations in 12-h windows (21--09 and 09--21\,UTC),
so a 12-h step from one of these times always crosses exactly one window boundary and the model never mixes steps
within a window with steps across one, as in GenCast\cite{price2025gencast}. The prognostic state has 82 variables: geopotential, temperature, specific humidity,
zonal and meridional wind and vertical velocity on 13 pressure levels (50, 100, 150, 200, 250, 300, 400, 500, 600,
700, 850, 925 and 1,000\,hPa), and 2-m temperature (T2m), 10-m zonal and meridional wind (U10, V10) and mean
sea-level pressure (MSLP) (Extended Data Table~\ref{edt:vars}). The forcings are SST, the land--sea mask, surface geopotential and top-of-atmosphere
insolation at $t$, $t+6$\,h and $t+12$\,h. Inputs are standardized with per-variable means and standard deviations; the network predicts the 12-h tendency standardized by the standard deviation of 12-h tendencies. Training
uses 1979--2018, 2019 is held out for validation and every design decision, and 2020 and 2022 are test years. During
a forecast, the initial SST anomaly with respect to a daily climatology (1979--2018, 15-day running mean) is kept
constant and added to the climatology of each valid day.

\subsection*{Model}
\emph{Stems.} The 55 dynamical channels (geopotential, zonal and meridional wind and vertical velocity on 13 levels,
MSLP, U10 and V10) and the 33 thermodynamical and forcing channels (temperature and specific humidity on 13 levels,
T2m, SST, two static fields and three insolation fields) pass through separate stems, each with two 3\,$\times$\,3
convolutions with 96 channels on the 1.5$^\circ$ grid and a strided 3\,$\times$\,2 patch convolution to 61\,$\times$\,120
tokens. Padding is periodic in longitude; across a pole, rows are continued from the meridian 180$^\circ$ away with the
signs of the wind components reversed. Each token stream receives a learned projection of real spherical harmonics up
to degree 20 at the token centers.

\emph{Transformer.} Step~1 has six blocks of width 384 on the dynamical tokens. Step~2 concatenates the
thermodynamical tokens (width 384) and has six blocks of width 768. Every block uses pre-normalization (RMSNorm) with
adaptive scale, shift and gate from a lead-time embedding (AdaLN\cite{peebles2023dit}), global multi-head attention
(head dimension 64) with two-dimensional rotary position embedding\cite{su2024roformer}, in which half of each
head's rotary pairs encode latitude and half encode longitude with integer harmonics so that the encoding is exactly
periodic, and a SwiGLU feed-forward layer\cite{shazeer2020glu} (expansion 8/3). Dropout with rate 0.1 acts on the
attention output and on the hidden SwiGLU activations; stochastic depth (0.1) is used only in the deterministic
stage.

\emph{Decoder.} The tokens are resized bilinearly to the 121\,$\times$\,240 grid (periodic in longitude, poles on grid
rows), convolved, concatenated with the full-resolution stem features and passed through two 3\,$\times$\,3
convolutions with 384 channels and an output projection to the 82 tendencies. The model has \params{} million
parameters (stems 0.7\,M, step~1 14.2\,M, step~2 49.6\,M, decoder 4.0\,M). \model-S uses widths 192/384, 64 stem
channels and 256 decoder channels (\paramsS\,M).

\subsection*{Training}
All stages weight grid points by the cosine of latitude, pressure levels in proportion to pressure, T2m with 1 and
the other surface variables with 0.1, as in GraphCast\cite{lam2023graphcast}. Two-dimensional weights in the transformer
blocks are optimized with Muon\cite{liu2025muon} and all other parameters with AdamW\cite{loshchilov2019adamw}, under
one cosine schedule with linear warm-up; an exponential moving average of the weights (0.999) is kept. Computation
uses bfloat16.

\emph{Stage 1 (deterministic).} 30 epochs over 1979--2018 with the RMSE loss of Otter\cite{otter2026}, in which the
RMSE of each variable is computed over the batch and the grid before the variables are averaged with their weights;
global batch 16, peak learning rate $5\times10^{-4}$ (warm-up 1,000 steps, final $5\times10^{-6}$), weight decay 0.1,
data parallel on two A40 GPUs.

\emph{Stage 2 (probabilistic).} Five epochs with the fair CRPS (equation~\eqref{eq:crps}) over $M=2$ members that
differ only in their dropout masks; batch 8, peak learning rate $2\times10^{-4}$ (warm-up 500 steps), no stochastic
depth.

\emph{Stage 3 (rollout).} One epoch over 2007--2018 with the fair CRPS of the state error averaged over autoregressive
rollouts whose length grows from 2 to 6 steps (24--72\,h) during the epoch; two members, batch 2, activation
checkpointing per step, and weight 1.0 on U10 and V10. Three models start from the stage-2 weights with learning rates
$10^{-4}$ (two seeds) and $3\times10^{-5}$. Stages 1 and 2 take 19.9 and 6.6 wall-clock hours on two A40s; each rollout run
takes 4.1\,h on one A40, so the three stages use 39.8, 13.2 and 12.3 GPU-hours.

\subsection*{Ensemble generation}
Dropout layers stay in training mode during forecasts and all other layers in evaluation mode. Each member draws
independent dropout masks at every step, starting from the same ERA5 state. The three stage-3 models contribute 17
members each (51 members). The power spectra and the WB2 cross-check use six per model (18 members). No initial-state perturbation is applied.

\subsection*{Verification}
All metrics follow WB2\cite{rasp2024wb2}. For variable $v$, lead time $\tau$ and initial times $t=1,\dots,T$, grid
points $i$ carry weights $w_i=\cos\phi_i/\overline{\cos\phi}$. With members $x^{(m)}$, $m=1,\dots,M$, and truth $y$,
the fair CRPS\cite{zamo2018estimation} is
\begin{equation}
\mathrm{CRPS}=\frac{1}{T}\sum_{t}\frac{1}{N}\sum_i w_i\left[\frac{1}{M}\sum_{m}\left|x^{(m)}_{t,i}-y_{t,i}\right|
-\frac{1}{2M(M-1)}\sum_{m,m'}\left|x^{(m)}_{t,i}-x^{(m')}_{t,i}\right|\right],
\label{eq:crps}
\end{equation}
an unbiased estimate of the CRPS of the distribution from which the members are drawn, so that ensembles of different
sizes can be compared. It is the estimator of the WB2 code, so \model{} and every reference are scored identically,
and it is also used by FGN, ArchesWeatherGen and U-Cast\cite{alet2025fgn,couairon2024archesgen,ucast2026}. The
traditional estimator, used in the GenCast paper\cite{price2025gencast}, adds the spread term divided by $M$ and so
penalizes small ensembles; with it, the headline skill of the 51-member \model{} against each reference changes by at
most \tradMaxShift{} percentage points (Supplementary Table~\ref{st:trad}). CRPS skill against a reference is $100\,(1-\mathrm{CRPS}/\mathrm{CRPS}_\mathrm{ref})$ for each
variable and lead time. The ensemble-mean squared error and the ensemble variance are
$\mathrm{MSE}=\frac{1}{T}\sum_t\frac{1}{N}\sum_i w_i(\bar x_{t,i}-y_{t,i})^2$ and
$\mathrm{Var}=\frac{1}{T}\sum_t\frac{1}{N}\sum_i w_i s^2_{t,i}$, with $s^2$ the unbiased member variance. We report the
ensemble-mean RMSE corrected for ensemble size, $\sqrt{\mathrm{MSE}-\mathrm{Var}/M}$, and the spread--skill ratio
$\sqrt{\mathrm{Var}/(\mathrm{MSE}-\mathrm{Var}/M)}$, which is 1 for a perfectly calibrated ensemble of any
size\cite{fortin2014spread}. The ACC is computed for each initial time from anomalies with respect to the WB2 ERA5
climatology (1990--2017) and then averaged. Scorecards cover Z, T, Q, U and V at 500, 700 and 850\,hPa and T2m, MSLP,
U10 and V10 at lead times of 0.5--15 days (570 targets). Zonal power spectra are computed along latitude circles
between 30$^\circ$ and 60$^\circ$ in both hemispheres and averaged over members and initial times. Brier scores are
computed for exceeding the monthly 95th percentile and falling below the monthly 5th percentile of ERA5 (1990--2019,
per grid point) and compared with the climatological forecast of 5\%. Our implementation reproduces the WB2 functions
(\texttt{weatherbench2.metrics}) on the same forecasts with a largest relative difference of
$7.7\times10^{-9}$ (Supplementary Table~\ref{st:wb2}).

\subsection*{Baselines}
Reference forecasts are taken from WB2 on the 1.5$^\circ$ grid: IFS ENS (50 members) and IFS HRES, initialized from
the operational analysis; GenCast (56 members) initialized from ERA5 (2020); GenCast operational, initialized from the
operational analysis (2020 and 2022); FGN operational (64 members, 2022); NeuralGCM ENS (50 members, 1.4$^\circ$,
2020); and ArchesWeatherGen (2020). Member counts are those of the WB2 forecast archives. All are verified against ERA5 with the same code. U-Cast values are taken from its paper\cite{ucast2026}. For the
case studies, IFS ENS and FGN forecasts are read from the WB2 forecast archives at 1.5$^\circ$. Tropical-cyclone
centers are tracked in every member as the MSLP minimum within 6$^\circ$ of the previous position, starting from the
ERA5 analysis, until the minimum exceeds 1,004\,hPa.
Training costs and inference times of other models are those reported in the U-Cast paper\cite{ucast2026}.

\subsection*{Spread-source analysis}
\emph{Activations.} We recorded the output of the last transformer block (step~2, block~6; 768 dimensions) of one
stage-3 model with dropout off on 1,200 states spread evenly over 2007--2018, keeping 1,024 random tokens per state
(1.2 million tokens). One dimension carries 59\% of the activation variance and five carry 95\%, so we standardized
each dimension with its training mean and standard deviation.

\emph{KAN-SAE.} We use the KAN-SAE\cite{cheon2026kansae} with its learned per-feature B-spline gains on the
decoder side. It encodes a token $h$ as $f=\mathrm{TopK}_{32}\big(\mathrm{ReLU}(W_\mathrm{enc}(h-b))
+b_\mathrm{enc}\big)$ with 2,048 features and decodes $\hat h=W_\mathrm{dec}\,\varphi(f)+b$, where $\varphi$
applies to each feature a learned degree-1 B-spline gain on 8 grid intervals, extended linearly beyond the
grid\cite{liu2024kan}. It is trained for 60 epochs with Adam
(learning rate $3\times10^{-4}$, ten times larger for the spline coefficients, cosine schedule, decoder columns
renormalized after each step) and an auxiliary loss in which the 256 most active dead features reconstruct the
residual (AuxK, coefficient 1/32)\cite{gao2024scaling}. On 5\% held-out tokens it explains \saeEv{} of the variance
with \saeDead\% dead features.

\emph{Spread experiments.} From every sixth 12-hourly initial time of 2019 (\nSaeInits{} times), one model runs eight
dropout members for six steps. At the first step we encode the activations of every member and token and compute the
member variance of each feature; its sum over features is the injected variance. Ensemble spreads of Z500, T850, T2m
and U10 are averaged onto the token grid; spread growth is the ratio of the geometric-mean spread at day 3 to that at
12\,h. The Eady growth rate is $\sigma=0.31\,f\,|\partial \mathbf{u}/\partial z|/N$ for the 850--500-hPa layer of
the ERA5 initial state\cite{eady1949long,lindzen1980simple}, with $f$ the Coriolis parameter and $N$ the
Brunt--V\"ais\"al\"a frequency. Spatial correlations are weighted by the cosine of latitude. For the out-of-sample
prediction, the target is the day-3 log spread minus its token climatology from the fitting times; odd initial times
are used for fitting and even ones for testing; predictors are standardized and fitted by ridge regression with the
penalty chosen on an inner split of the fitting times; the KAN-SAE predictors are the activation and dropout spread of
the features most correlated with the target on the fitting times.

\subsection*{Computing}
Training and evaluation ran on one server with three NVIDIA A40 GPUs (48\,GB, no NVLink), using at most two GPUs for
\model{}. Inference times are for one model and one member with bfloat16 after warm-up.

\section*{Data availability}
ERA5 and all baseline forecasts are available from WeatherBench~2 (\url{https://weatherbench2.readthedocs.io}).
\model{} forecasts for 2020 and 2022 will be deposited in a public archive on publication.

\section*{Code availability}
Code and trained weights will be released on publication.

\section*{Author contributions}
M.C. designed the study, developed the model, ran the experiments and wrote the paper.

\section*{Competing interests}
The author declares no competing interests.

% ---------------------------------------------------------------------------------------------------------------
\clearpage
\section*{Extended Data}
\setcounter{figure}{0}
\setcounter{table}{0}
\captionsetup[figure]{name={Extended Data Fig.}}
\captionsetup[table]{name={Extended Data Table}}

\begin{figure}[htbp]
\centering
\includegraphics[width=\textwidth]{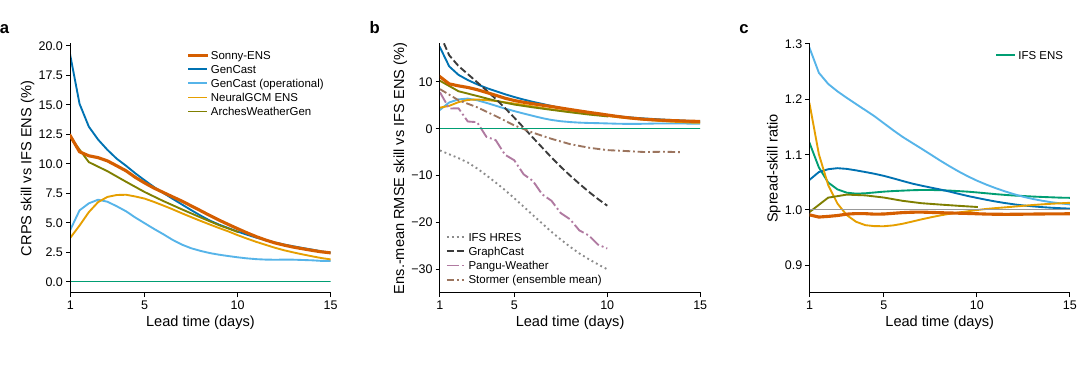}
\caption{\textbf{| Summary against IFS ENS on 2020.} Averages over the five upper-air variables used to compare
ArchesWeatherGen and U-Cast (Z500, T850, Q700, U850 and V850) of \textbf{a}, the CRPS skill of every WB2 ensemble;
\textbf{b}, the ensemble-mean RMSE skill (plain mean of each ensemble), with the deterministic IFS HRES, GraphCast and
Pangu-Weather and the Stormer ensemble mean (dashed); and \textbf{c}, the spread--skill ratio corrected for ensemble
size. Skill is $100\,(1-\mathrm{score}/\mathrm{score}_\mathrm{ENS})$ for each variable and lead time, then averaged.
\model{} has 51 members.}
\label{edf:summary}
\end{figure}

\begin{figure}[htbp]
\centering
\includegraphics[width=\textwidth]{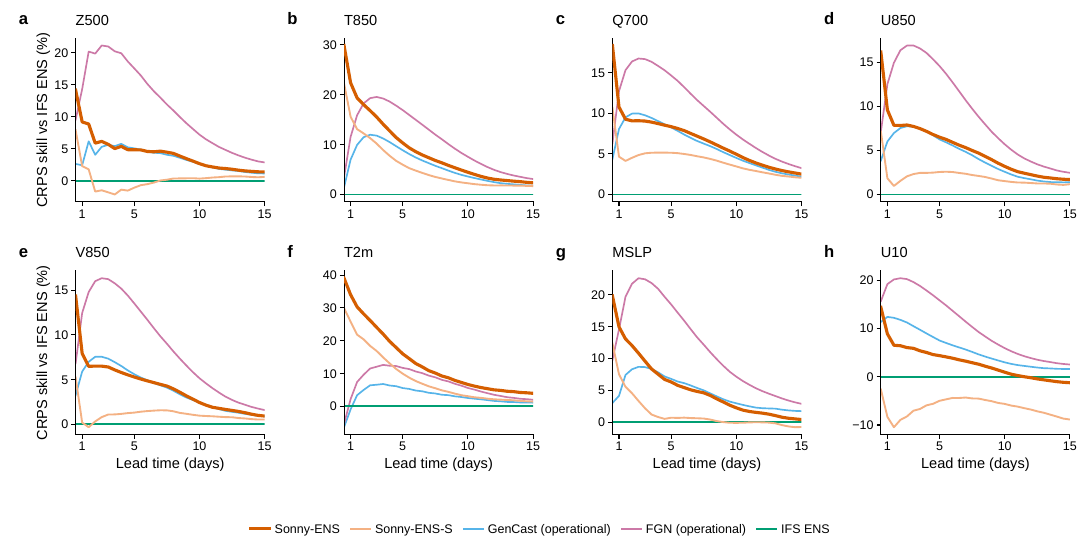}
\caption{\textbf{| Skill against ENS on 2022.} CRPS skill relative to IFS ENS for \model, GenCast (operational) and
FGN (operational).}
\label{edf:skill22}
\end{figure}

\begin{figure}[htbp]
\centering
\includegraphics[width=\textwidth]{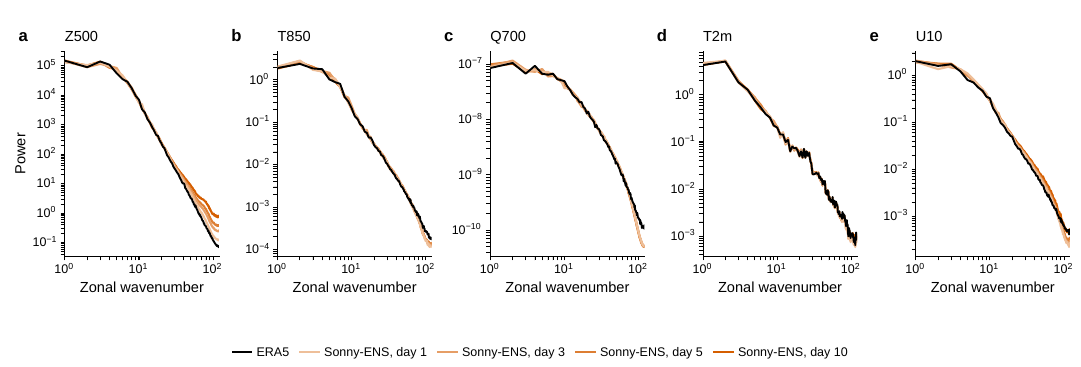}
\caption{\textbf{| Power spectra.} Zonal power spectra of \model{} members and ERA5 (30--60$^\circ$ latitude) at
days 1, 3, 5 and 10 in 2022 (23 initial times, 18 members).}
\label{edf:spec}
\end{figure}

\begin{figure}[htbp]
\centering
\includegraphics[width=\textwidth]{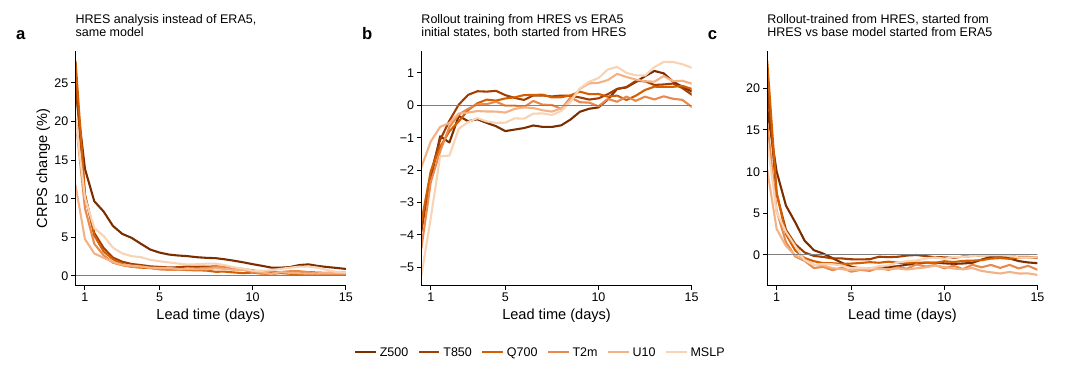}
\caption{\textbf{| The source of the initial state matters mostly at day 1.} CRPS change on 2019 (8 members,
\model-S). \textbf{a}, The same model started from the HRES operational analysis instead of ERA5 (truth ERA5):
+13.9\% (Z500), +10.4\% (T850) and +8.8\% (T2m) at day 1, below 1.5\% by day 10. \textbf{b}, Rollout training on
HRES initial states against training on ERA5 states, both started from HRES. \textbf{c}, The HRES-trained model
started from HRES against the base model started from ERA5.}
\label{edf:init}
\end{figure}

\begin{figure}[htbp]
\centering
\includegraphics[width=\textwidth]{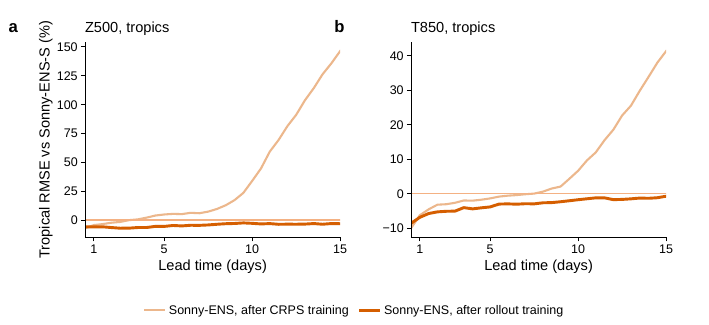}
\caption{\textbf{| Rollout training removes the tropical drift of CRPS training.} Tropical ensemble-mean RMSE on 2019
relative to \model-S after rollout training. After CRPS training alone, tropical Z500 RMSE is 34\% higher at day 10
and 146\% higher at day 15; after rollout training it is 2.8\% and 3.1\% lower.}
\label{edf:drift}
\end{figure}

\begin{figure}[htbp]
\centering
\includegraphics[width=\textwidth]{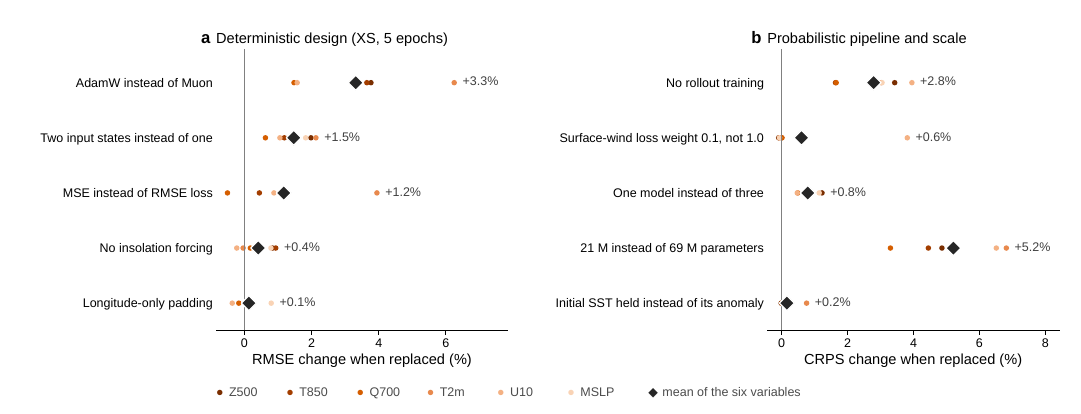}
\caption{\textbf{| What each design choice is worth.} Change on 2019 when the adopted choice is replaced (positive:
the alternative is worse); diamonds, mean of six variables (Z500, T850, Q700, T2m, U10 and MSLP) over lead days
1--10 or 1--15; dots, each variable. \textbf{a}, Deterministic design (smallest model, 5 epochs, RMSE loss).
\textbf{b}, Probabilistic pipeline and scale (CRPS; the size row compares single models with the same recipe and
seed).}
\label{edf:abl}
\end{figure}

\begin{figure}[htbp]
\centering
\includegraphics[width=\textwidth]{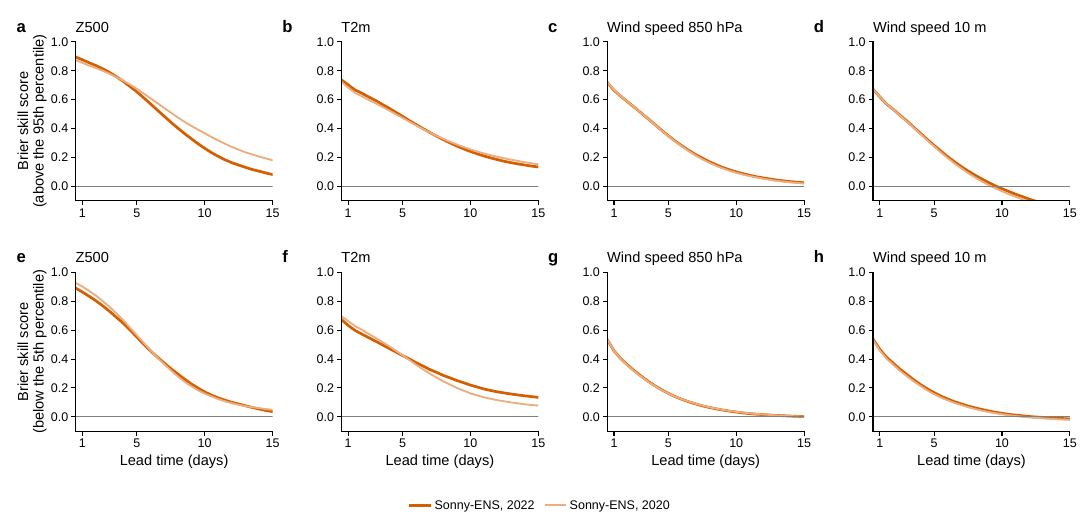}
\caption{\textbf{| Extreme events.} Brier skill score of the 51-member \model{} for exceeding the monthly ERA5
(1990--2019) 95th percentile (top) and falling below the 5th (bottom), per grid point, against the climatological 5\%
forecast, for Z500, T2m and wind speed at 850\,hPa and 10\,m, on 2022 (\initsTwentyTwo{} initial times) and 2020
(\initsTwenty).}
\label{edf:ext}
\end{figure}

\begin{table}[htbp]
\centering
\caption{\textbf{| Variables.} All fields are ERA5 on the 1.5$^\circ$ grid (121\,$\times$\,240) every 12\,h; upper-air
variables on 13 pressure levels (50, 100, 150, 200, 250, 300, 400, 500, 600, 700, 850, 925 and 1,000\,hPa). Stem, the
convolutional stem that encodes the field (dynamics, 55 channels; thermodynamics, 33 channels). Loss weight, before
normalization to a mean of 1 over the 82 prognostic channels; $p/\bar p$, pressure of the level divided by the mean
of the 13 levels; all losses are also weighted by the cosine of latitude. Verified, levels scored in the scorecards
(headline variables: Z500, T850, Q700, U850, T2m, U10 and MSLP). $^{a}$0.1 in the deterministic and CRPS stages, 1.0
in rollout training. $^{b}$During forecasts the initial SST anomaly with respect to the 1979--2018 daily climatology is
kept and added to the climatology of each valid day. $^{c}$At $t$, $t+6$\,h and $t+12$\,h, computed from the solar
geometry as a fraction of the solar constant. Lead time is fixed at 12\,h.}
\label{edt:vars}
\footnotesize
\setlength{\tabcolsep}{3pt}
\begin{tabular}{llllllll}
\toprule
Type & Variable & Symbol & Channels & Units & Stem & Loss weight & Verified (hPa)\\
\midrule
\multicolumn{8}{l}{\textit{Prognostic variables (input and output; 82 channels)}}\\
Upper air & Geopotential & Z & 13 & m$^2$\,s$^{-2}$ & Dynamics & $p/\bar p$ & 500/700/850 \\
 & Temperature & T & 13 & K & Thermodynamics & $p/\bar p$ & 500/700/850 \\
 & Specific humidity & Q & 13 & kg\,kg$^{-1}$ & Thermodynamics & $p/\bar p$ & 500/700/850 \\
 & Zonal wind & U & 13 & m\,s$^{-1}$ & Dynamics & $p/\bar p$ & 500/700/850 \\
 & Meridional wind & V & 13 & m\,s$^{-1}$ & Dynamics & $p/\bar p$ & 500/700/850 \\
 & Vertical velocity & W & 13 & Pa\,s$^{-1}$ & Dynamics & $p/\bar p$ & -- \\
Surface & 2-m temperature & T2m & 1 & K & Thermodynamics & 1.0 & Yes \\
 & 10-m zonal wind & U10 & 1 & m\,s$^{-1}$ & Dynamics & 0.1 / 1.0$^{a}$ & Yes \\
 & 10-m meridional wind & V10 & 1 & m\,s$^{-1}$ & Dynamics & 0.1 / 1.0$^{a}$ & Yes \\
 & Mean sea-level pressure & MSLP & 1 & Pa & Dynamics & 0.1 & Yes \\
\multicolumn{8}{l}{\textit{Forcings (input only; 6 channels)}}\\
Boundary & Sea-surface temperature$^{b}$ & SST & 1 & K & Thermodynamics & -- & -- \\
Static & Land--sea mask & LSM & 1 & 0--1 & Thermodynamics & -- & -- \\
 & Surface geopotential & Z$_\mathrm{s}$ & 1 & m$^2$\,s$^{-2}$ & Thermodynamics & -- & -- \\
Solar & TOA insolation$^{c}$ & -- & 3 & 1 & Thermodynamics & -- & -- \\
\multicolumn{8}{l}{\textit{Conditioning}}\\
Time & Lead time & $\Delta t$ & -- & h & AdaLN & -- & -- \\
\bottomrule
\end{tabular}
\end{table}

\begin{table}[htbp]
\centering
\caption{\textbf{| Scores on 2020 at days 1, 3, 5 and 10.} \textbf{a}, CRPS. \textbf{b}, Ensemble-mean RMSE
(corrected for ensemble size; IFS HRES and climatology uncorrected). Lower is better; best ensemble in bold. Units:
Z500 (m$^2$\,s$^{-2}$), T850 and T2m (K), U10 (m\,s$^{-1}$), MSLP (Pa). \model: three models $\times$ 17 dropout
members, ERA5 initial states; \model-S: three models $\times$ six. References from WB2 (GenCast operational and IFS from operational analyses); U-Cast as
reported.}
\label{edt:scores}
\scriptsize
\setlength{\tabcolsep}{2.2pt}
\resizebox{\textwidth}{!}{%
\begin{tabular}{l*{20}{r}}
\toprule
 & \multicolumn{4}{c}{Z500} & \multicolumn{4}{c}{T850} & \multicolumn{4}{c}{T2m} & \multicolumn{4}{c}{U10}
 & \multicolumn{4}{c}{MSLP}\\
\cmidrule(lr){2-5}\cmidrule(lr){6-9}\cmidrule(lr){10-13}\cmidrule(lr){14-17}\cmidrule(lr){18-21}
Day & 1 & 3 & 5 & 10 & 1 & 3 & 5 & 10 & 1 & 3 & 5 & 10 & 1 & 3 & 5 & 10 & 1 & 3 & 5 & 10\\
\midrule
\multicolumn{21}{l}{\textbf{a} CRPS}\\
Sonny-ENS (51 members) & 21.6 & 55.3 & 109.8 & 254.6 & 0.295 & 0.464 & 0.690 & \textbf{1.279} & 0.269 & 0.379 & 0.511 & 0.845 & 0.384 & 0.642 & 0.956 & 1.582 & 26.9 & 59.9 & 112.2 & 241.2 \\
Sonny-ENS-S (18 members) & 23.2 & 59.7 & 116.6 & 261.5 & 0.322 & 0.493 & 0.723 & 1.307 & 0.303 & 0.414 & 0.546 & 0.876 & 0.457 & 0.731 & 1.051 & 1.689 & 29.3 & 65.0 & 119.4 & 248.3 \\
U-Cast (reported) & \textbf{19.6} & \textbf{53.5} & -- & \textbf{253.0} & -- & -- & -- & -- & -- & -- & -- & -- & 0.345 & \textbf{0.600} & -- & \textbf{1.540} & -- & -- & -- & -- \\
GenCast & 20.2 & 54.3 & \textbf{108.3} & 253.6 & \textbf{0.269} & \textbf{0.459} & \textbf{0.687} & 1.281 & \textbf{0.209} & \textbf{0.343} & \textbf{0.482} & \textbf{0.828} & \textbf{0.332} & 0.604 & \textbf{0.921} & 1.547 & \textbf{24.6} & \textbf{57.7} & \textbf{109.6} & \textbf{239.0} \\
GenCast (operational) & 24.9 & 58.6 & 114.7 & 262.6 & 0.341 & 0.494 & 0.724 & 1.316 & 0.400 & 0.477 & 0.596 & 0.908 & 0.377 & 0.631 & 0.956 & 1.579 & 30.8 & 63.2 & 116.8 & 246.9 \\
NeuralGCM ENS & 22.9 & 54.9 & 108.4 & 253.5 & 0.333 & 0.497 & 0.715 & 1.293 & -- & -- & -- & -- & -- & -- & -- & -- & -- & -- & -- & -- \\
ArchesWeatherGen & 21.2 & 55.8 & 110.8 & 254.1 & 0.297 & 0.470 & 0.699 & 1.289 & 0.245 & 0.366 & 0.506 & 0.847 & 0.370 & 0.625 & 0.935 & 1.546 & 25.8 & 59.4 & 112.0 & 239.5 \\
IFS ENS & 24.1 & 59.3 & 116.3 & 263.8 & 0.370 & 0.550 & 0.775 & 1.343 & 0.396 & 0.493 & 0.610 & 0.911 & 0.415 & 0.679 & 1.000 & 1.602 & 32.1 & 65.6 & 118.7 & 247.0 \\
\midrule
\multicolumn{21}{l}{\textbf{b} Ensemble-mean RMSE}\\
Sonny-ENS (51 members) & 41.6 & 126.1 & 266.0 & 602.9 & 0.587 & 0.960 & 1.488 & 2.718 & 0.608 & 0.867 & 1.210 & 2.029 & 0.748 & 1.358 & 2.075 & 3.301 & 54.0 & 136.0 & 268.7 & 561.3 \\
GenCast & \textbf{39.0} & \textbf{122.6} & \textbf{260.4} & 601.5 & \textbf{0.537} & \textbf{0.948} & \textbf{1.473} & \textbf{2.715} & \textbf{0.490} & \textbf{0.795} & \textbf{1.158} & \textbf{2.005} & \textbf{0.661} & \textbf{1.308} & \textbf{2.033} & 3.275 & \textbf{48.2} & \textbf{130.2} & \textbf{262.1} & \textbf{559.2} \\
GenCast (operational) & 46.3 & 130.3 & 274.0 & 620.0 & 0.682 & 1.010 & 1.539 & 2.775 & 0.842 & 1.028 & 1.340 & 2.129 & 0.740 & 1.347 & 2.089 & 3.325 & 59.3 & 138.8 & 275.1 & 574.9 \\
NeuralGCM ENS & 43.4 & 124.5 & 263.0 & \textbf{600.4} & 0.653 & 1.017 & 1.520 & 2.729 & -- & -- & -- & -- & -- & -- & -- & -- & -- & -- & -- & -- \\
ArchesWeatherGen & 41.5 & 129.0 & 270.0 & 604.5 & 0.590 & 0.973 & 1.504 & 2.729 & 0.538 & 0.826 & 1.196 & 2.030 & 0.736 & 1.353 & 2.065 & \textbf{3.274} & 51.2 & 136.0 & 270.0 & 561.7 \\
IFS ENS & 45.3 & 131.9 & 275.7 & 617.1 & 0.713 & 1.095 & 1.607 & 2.793 & 0.837 & 1.063 & 1.376 & 2.134 & 0.811 & 1.440 & 2.162 & 3.348 & 63.7 & 144.6 & 278.2 & 571.1 \\
IFS HRES & 47.2 & 137.2 & 305.5 & 802.7 & 0.730 & 1.176 & 1.836 & 3.654 & 0.885 & 1.130 & 1.521 & 2.693 & 0.865 & 1.608 & 2.568 & 4.481 & 62.6 & 149.8 & 309.5 & 748.7 \\
Climatology & 819.4 & 820.1 & 820.9 & 821.8 & 3.444 & 3.442 & 3.441 & 3.448 & 2.610 & 2.610 & 2.610 & 2.613 & 3.824 & 3.825 & 3.828 & 3.829 & 724.1 & 724.7 & 725.8 & 726.3 \\
\bottomrule
\end{tabular}}
\end{table}

\begin{table}[htbp]
\centering
\caption{\textbf{| Model size, training cost and inference speed.} \model: deterministic stage 30 epochs and CRPS
stage 5 epochs on two GPUs, rollout stage one epoch per model on one GPU, measured wall time; inference, one model and
one member for 60 12-h steps (30 days) in bfloat16 after warm-up. Other rows as reported in the U-Cast paper.
A100-equivalent days use the factors of ArchesWeatherGen and U-Cast\cite{couairon2024archesgen,ucast2026} (V100 0.5,
H100 2, TPUv5e 3, H200 2.5) and 0.5 for the A40, which has half the dense bfloat16 throughput of an A100. $^{a}$TPUv5
and TPUv5p counted as TPUv5e.}
\label{edt:cost}
\small
\resizebox{\textwidth}{!}{%
\begin{tabular}{llllll}
\toprule
Model & Resolution & Parameters & Training (accelerator-days) & A100-equivalent days & Inference, 1 member, 60 steps\\
\midrule
\model & 1.5$^\circ$ & 68.9\,M $\times$ 3 & 2.7 (A40) & 1.4 & 3.2 s (A40)\\
\model-S & 1.5$^\circ$ & 20.8\,M $\times$ 3 & 1.3 (A40) & 0.6 & 1.7 s (A40)\\
U-Cast & 1.5$^\circ$ & 895\,M $\times$ 4 & 11.8 (H200) & 29.5 & 3.6 s (A100)\\
ArchesWeatherGen & 1.5$^\circ$ & -- & 45 (V100) & 22.5 & 3.5 min (A100)\\
NeuralGCM ENS & 1.4$^\circ$ & -- & 1,280 (TPUv5e) & 3,840 & 18.6 s (TPUv4)\\
GenCast & 0.25$^\circ$ & -- & 160 (TPUv5) & 480$^{a}$ & $\sim$16 min (TPUv5)\\
FGN & 0.25$^\circ$ & -- & 490 (TPUv5p/6e) & 1,470$^{a}$ & $\sim$1 min (TPUv5p)\\
AIFS-CRPS & $\sim$0.25$^\circ$ & -- & 896 (H100) & 1,792 & $\sim$4 min (A100)\\
IFS ENS & $\sim$0.2$^\circ$ & -- & -- & -- & 1 h (96 CPUs)\\
\bottomrule
\end{tabular}}
\end{table}

% ---------------------------------------------------------------------------------------------------------------
\clearpage
\section*{Supplementary Information}
\setcounter{figure}{0}
\setcounter{table}{0}
\captionsetup[figure]{name={Supplementary Fig.}}
\captionsetup[table]{name={Supplementary Table}}

\begin{figure}[h]
\centering
\includegraphics[width=\textwidth]{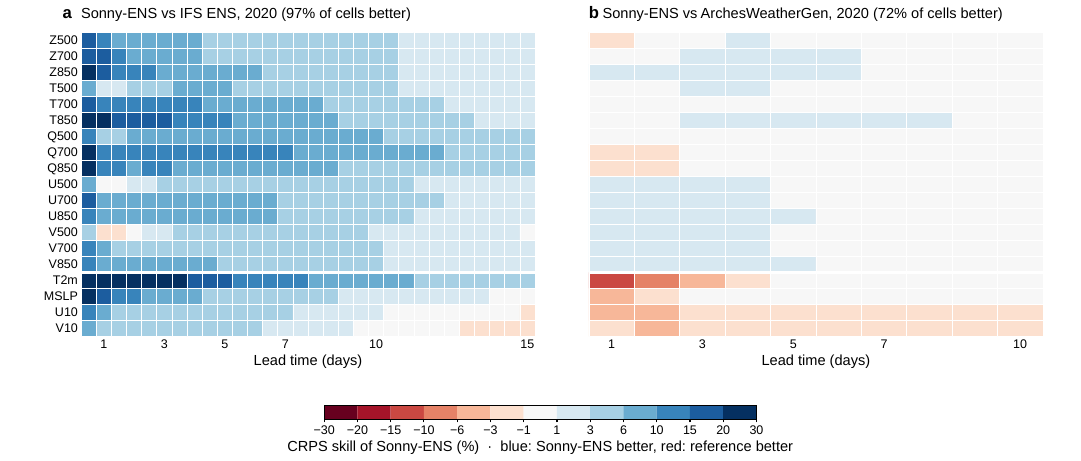}
\caption{\textbf{| Scorecards on 2020.} CRPS skill of \model{} against \textbf{a}, IFS ENS and \textbf{b},
ArchesWeatherGen, which runs at 1.5$^\circ$ like \model{}, by variable and lead time (\initsTwenty{} initial times).
ArchesWeatherGen is archived in WB2 at daily lead times up to 10 days.}
\label{sf:card20}
\end{figure}

\begin{figure}[htbp]
\centering
\includegraphics[width=\textwidth]{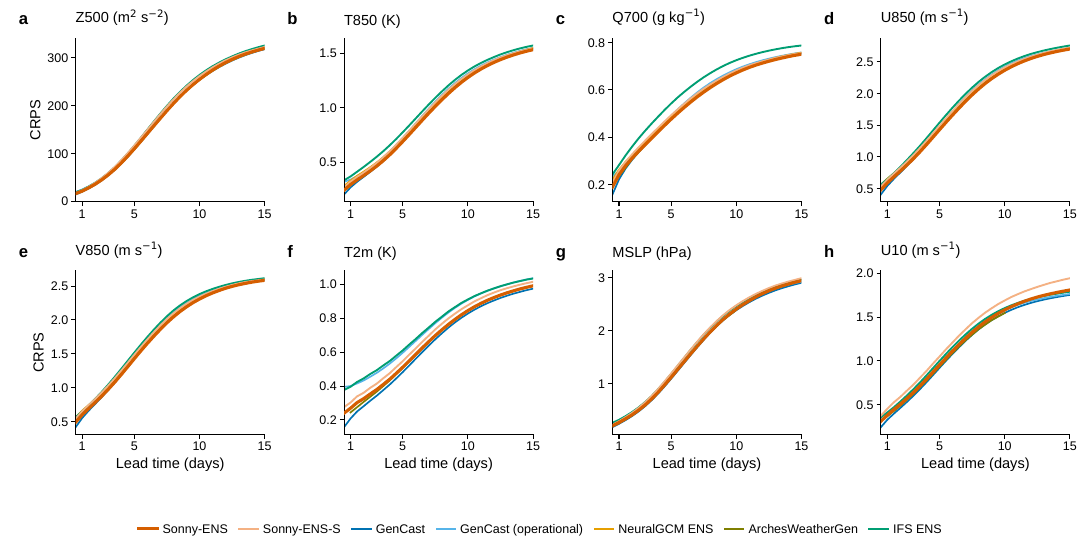}
\caption{\textbf{| Absolute CRPS on 2020} in physical units for all WB2 ensembles.}
\label{sf:abs}
\end{figure}

\begin{figure}[htbp]
\centering
\includegraphics[width=\textwidth]{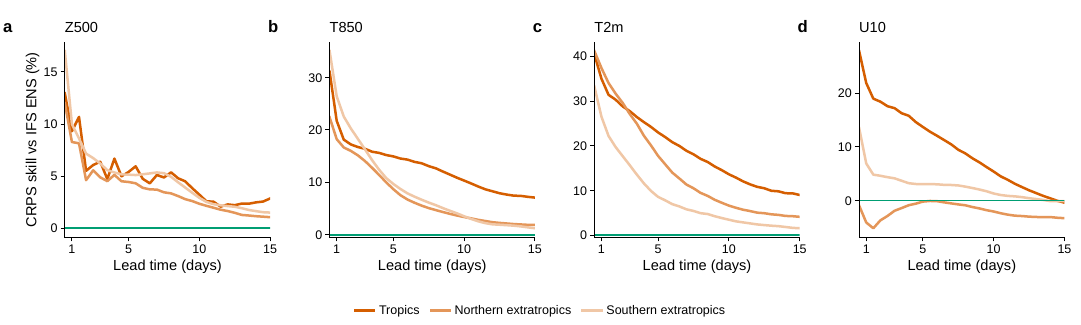}
\caption{\textbf{| Skill against ENS by latitude band}, 2022 (tropics, $|\phi|\le20^\circ$; extratropics poleward of
20$^\circ$).}
\label{sf:reg}
\end{figure}

\begin{figure}[h]
\centering
\includegraphics[width=\textwidth]{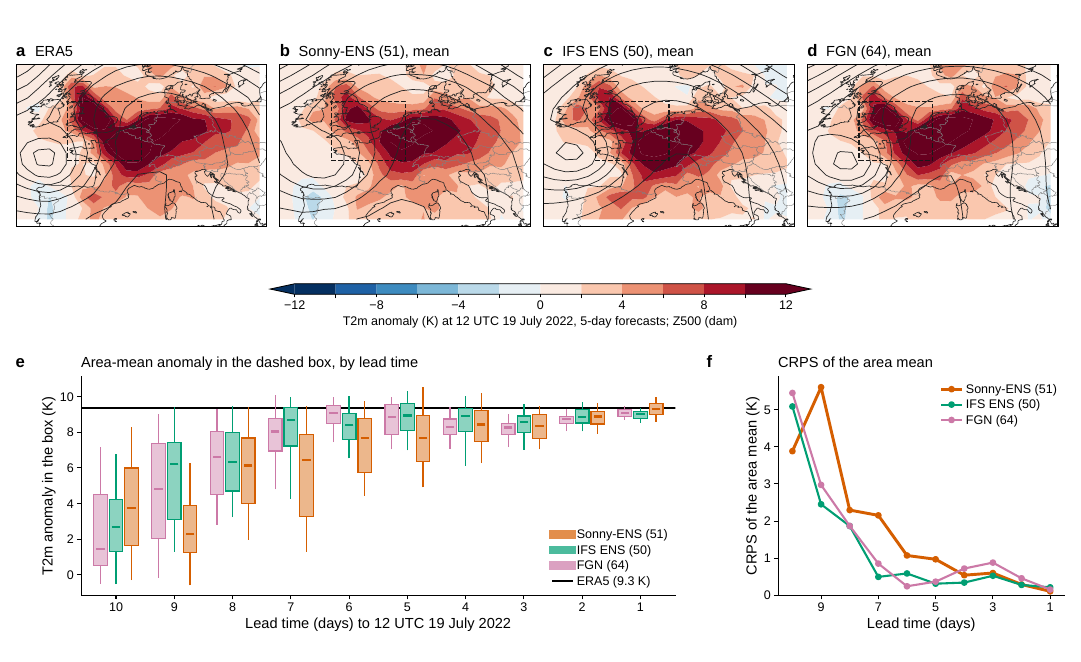}
\caption{\textbf{| The 2022 heatwave in three ensembles.} \textbf{a--d}, T2m anomaly at 12\,UTC 19 July 2022 from
forecasts started 5 days earlier: ERA5 and the ensemble means of \model{} (51 members), IFS ENS (50) and FGN (64);
contours, Z500 (dam). \textbf{e}, Area-mean anomaly in the dashed box from forecasts started 1--10 days earlier
(boxes, quartiles of the members; whiskers, 5--95\%; bars, median) and ERA5 (line). \textbf{f}, CRPS of the area
mean (fair estimator). At the 5-day lead the ensemble-mean area anomaly is 7.7\,K (\model), 8.8\,K (IFS ENS) and
8.7\,K (FGN), against 9.35\,K in ERA5; averaged over lead times of 1--5 days the CRPS is 0.50, 0.33 and 0.52\,K.
\model{} starts from ERA5, IFS ENS and FGN from the operational analysis.}
\label{sf:hw}
\end{figure}

\begin{table}[htbp]
\centering
\caption{\textbf{| Agreement with the WB2 metric code.} Largest relative difference over 20 lead times between our
CRPS, ensemble-mean MSE and ensemble variance and the official WB2 functions on the same 18-member forecasts.}
\small
\begin{tabular}{llrrr}
\toprule
Case & Variable & CRPS & Ens.-mean MSE & Ens. variance\\
\midrule
Heatwave 2022 & Z500 & $5.1\times10^{-9}$ & $7.5\times10^{-9}$ & $6.1\times10^{-9}$\\
 & T850 & $3.9\times10^{-9}$ & $4.7\times10^{-9}$ & $4.0\times10^{-9}$\\
 & Q700 & $3.4\times10^{-9}$ & $3.3\times10^{-9}$ & $3.1\times10^{-9}$\\
 & T2m & $3.7\times10^{-9}$ & $5.0\times10^{-9}$ & $3.3\times10^{-9}$\\
 & MSLP & $5.4\times10^{-9}$ & $7.7\times10^{-9}$ & $6.9\times10^{-9}$\\
 & U10 & $4.5\times10^{-9}$ & $7.1\times10^{-9}$ & $5.8\times10^{-9}$\\
 & V10 & $4.3\times10^{-9}$ & $6.4\times10^{-9}$ & $5.5\times10^{-9}$\\
 & U850 & $4.8\times10^{-9}$ & $7.2\times10^{-9}$ & $6.4\times10^{-9}$\\
 & V850 & $4.7\times10^{-9}$ & $7.3\times10^{-9}$ & $6.2\times10^{-9}$\\
Hinnamnor 2022 & Z500 & $5.0\times10^{-9}$ & $7.2\times10^{-9}$ & $7.1\times10^{-9}$\\
 & T850 & $4.1\times10^{-9}$ & $5.3\times10^{-9}$ & $5.1\times10^{-9}$\\
 & Q700 & $3.7\times10^{-9}$ & $4.6\times10^{-9}$ & $3.2\times10^{-9}$\\
 & T2m & $3.6\times10^{-9}$ & $5.6\times10^{-9}$ & $3.4\times10^{-9}$\\
 & MSLP & $4.8\times10^{-9}$ & $6.8\times10^{-9}$ & $6.1\times10^{-9}$\\
 & U10 & $3.8\times10^{-9}$ & $6.2\times10^{-9}$ & $4.5\times10^{-9}$\\
 & V10 & $3.3\times10^{-9}$ & $5.1\times10^{-9}$ & $5.4\times10^{-9}$\\
 & U850 & $4.0\times10^{-9}$ & $5.8\times10^{-9}$ & $5.4\times10^{-9}$\\
 & V850 & $4.2\times10^{-9}$ & $6.9\times10^{-9}$ & $5.5\times10^{-9}$\\
\bottomrule
\end{tabular}
\label{st:wb2}
\end{table}

\begin{table}[h]
\centering
\caption{\textbf{| Headline CRPS skill with the fair and the traditional estimator.} Mean CRPS skill (\%) of the
51-member \model{} (three models $\times$ 17 dropout members) over the seven headline variables at lead times of 1--10
days; in parentheses, the share (\%) of the 133 targets on which \model{} is better. The traditional estimator adds
$\lambda_2/M$ to the fair CRPS, with $\lambda_2$ the fair spread term of equation~\eqref{eq:crps}. WB2 archives only
the fair CRPS of the references, so $\lambda_2$ is estimated from the ensemble spread $s$ as $\kappa s/\sqrt{\pi}$
(the Gaussian value for $\kappa=1$); $\kappa$ (\kappaLo--\kappaHi, by variable and lead time) is measured exactly on
\model{} forecasts of two 2022 cases and applied to every ensemble. The last column uses $\kappa=0.7$ and $1.0$ for
the reference.}
\label{st:trad}
\small
\begin{tabular}{llrrr}
\toprule
Year & Reference (members) & Fair & Traditional & Traditional, reference $\kappa=0.7$ / $1.0$\\
\midrule
2020 & IFS ENS (50) & $+8.8$ (100) & $+9.0$ (100) & $+8.7$ / $+9.4$ \\
2020 & GenCast (56) & $-2.8$ (24) & $-2.9$ (23) & $-3.1$ / $-2.4$ \\
2020 & NeuralGCM ENS (50) & $+2.4$ (79) & $+2.4$ (79) & $+2.1$ / $+2.8$ \\
2022 & IFS ENS (50) & $+8.3$ (100) & $+8.3$ (100) & $+8.1$ / $+8.8$ \\
2022 & FGN (operational) (64) & $-6.3$ (17) & $-6.8$ (16) & $-7.0$ / $-6.5$ \\
\bottomrule
\end{tabular}
\end{table}

\end{document}